\RequirePackage[T1]{fontenc}
\documentclass[letterpaper,10pt,conference]{IEEEtran}

\IEEEoverridecommandlockouts

\usepackage{amsmath,amssymb}
\usepackage{array}
\usepackage{booktabs}
\usepackage{capt-of}
\usepackage{cite}
\usepackage{graphicx}
\usepackage{multirow}
\usepackage{url}
\usepackage{xcolor}
\usepackage[hidelinks]{hyperref}
\definecolor{projectlink}{RGB}{15,35,110}
\hypersetup{
  pdftitle={ClothDojo: Learned Data Generation and Benchmarking for Bimanual Garment Flattening and Folding},
  pdfauthor={Zilong Huang, Zipeng Ye, Caicheng Wang, Yuchen Xie, Jiaxiang Luo}
}

\newcolumntype{T}{>{\centering\arraybackslash}m{0.43in}}

\title{\fontsize{22}{25}\selectfont ClothDojo: Learned Data Generation and Benchmarking\\
for Bimanual Garment Flattening and Folding}

\author{\IEEEauthorblockN{Zilong Huang$^{1,2}$, Zipeng Ye$^{2,\dagger}$, Caicheng Wang$^{2}$,
Yuchen Xie$^{2}$, Jiaxiang Luo$^{1,*}$}
\IEEEauthorblockA{$^{1}$South China University of Technology, $^{2}$Meituan\\
\href{https://clothdojo.github.io/}{\textcolor{projectlink}{https://clothdojo.github.io/}}}%
\thanks{$^{\dagger}$Project lead. \quad $^{*}$Corresponding author.}
}

\IEEEaftertitletext{%
  \begin{minipage}{\textwidth}
    \centering
  \includegraphics[width=\textwidth]{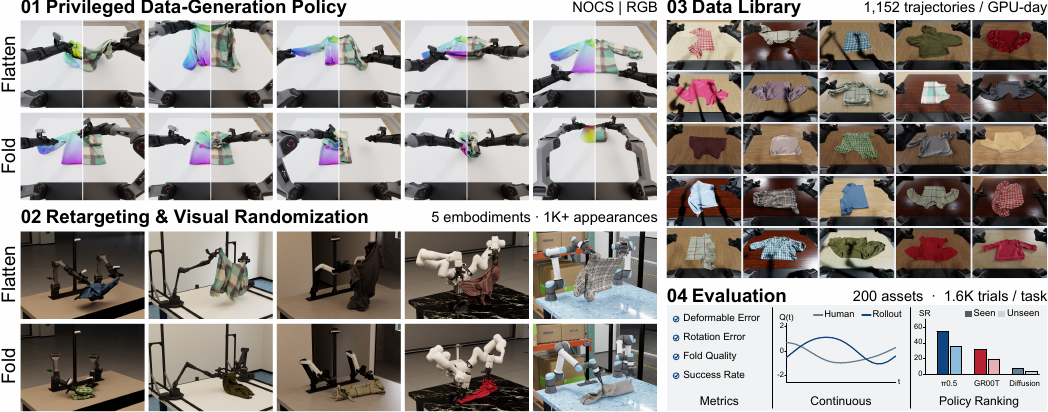}
  \captionof{figure}{\textbf{ClothDojo connects data generation, data
  expansion, and closed-loop evaluation.} (01) A data-generation
  policy with privileged visual observations generates flattening and
  folding trajectories, shown with normalized object coordinate space
  (NOCS) and standard RGB appearances side by side. (02) Retargeting and
  visual randomization expand embodiment and appearance coverage. (03) On a
  single NVIDIA GeForce RTX 5090 D, the pipeline executes approximately
  1,152 trajectories per GPU-day, excluding RGB re-rendering. (04)
  Evaluation summarizes discrete success criteria, continuous task metrics,
  and policy rankings.}
  \label{fig:framework}
  \end{minipage}%
}

\begin{document}

\maketitle
\raggedbottom
\thispagestyle{empty}
\pagestyle{empty}

\begin{abstract}
Garment flattening and folding require robots to identify garment structure
during manipulation and adapt their actions as the cloth deforms. We present
ClothDojo, a simulation platform that generates adaptive manipulation trajectories at scale
across diverse garments and initial states and evaluates policies in closed loop. ClothDojo encodes each garment's canonical mesh coordinates as an
NOCS texture, providing canonical surface-position cues through the visual
interface of a pretrained vision-language-action model. We fine-tune the model
on human demonstrations and corrections to obtain a data-generation policy.
The policy autonomously generates simulated manipulation trajectories, and we
retain rollouts that satisfy task-specific success criteria. Each retained
trajectory is re-rendered under a randomly sampled appearance to provide RGB
observations for downstream policies that do not receive NOCS. Across both
tasks, the policy produces nearly 8,000 successful trajectories over hundreds
of garments. RGB policies trained on these generated trajectories outperform
those trained directly on the original human demonstrations and corrections
across all evaluated garment splits and appearance settings. Evaluations
across multiple policy models show that overall performance and appearance
robustness vary across models and tasks, while generalization to unseen garments
remains a common challenge.
\end{abstract}

\begin{figure*}[t]
  \centering
  \includegraphics[width=\textwidth]{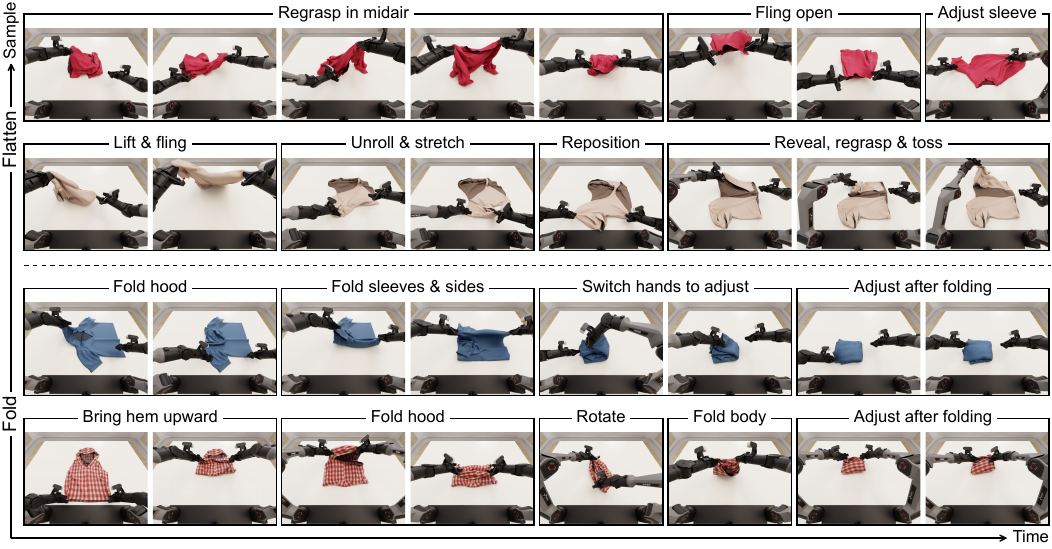}
  \caption{\textbf{Autonomous policies generate long-horizon behaviors with
  varied action transitions and adjustments during execution.} Flattening
  trajectories combine midair regrasping, dynamic unfolding, and local
  adjustments; folding includes hand switching, garment rotation, and
  post-fold adjustments. Each row shows keyframes from an independent rollout,
  ordered from left to right with nonuniform time intervals. Box labels
  describe observed behaviors; the images are re-rendered from recorded
  simulation states.}
  \label{fig:folding-trajectories}
\end{figure*}

\section{Introduction}
\label{sec:introduction}

Garment flattening and folding require visual understanding and bimanual
control throughout long sequences of interaction. Garments differ in structure
and appearance. Wrinkles and self-occlusion hide relevant parts, and each
contact changes the cloth configuration and the conditions for subsequent
actions. Policies must continually adjust to the outcomes of their actions.
Training across garments therefore requires continuous trajectories covering
many initial states. Collecting these trajectories on real robots is
time-consuming and requires repeatedly setting up complex configurations that
are difficult to reproduce. Differences in robots and scenes further limit
the reuse of data and experiments, making training and evaluation at scale
costly.

Simulation allows initial states to be reused and rollouts to run in batches,
but successful trajectories must still be designed or collected. Task-specific
primitives, keypoint planners, and prescribed scripts can execute fixed
procedures at scale, but cannot readily capture the state-dependent action
transitions in long-horizon garment manipulation. Fig.~\ref{fig:folding-trajectories}
illustrates such behaviors: flattening involves midair regrasping, dynamic
unfolding, and local adjustment, while folding involves hand switching,
rotation, and post-fold refinement. Their timing, arm assignment, and subsequent
adjustments depend on the cloth state produced by preceding actions and are
difficult to prescribe as a uniform procedure. Human teleoperation naturally
produces such trajectories, but its cost scales with the number collected. We
therefore study how to extend adaptive garment manipulation to more garments
and initial states.

Real-robot laundry systems show that demonstrations, human corrections, and
autonomous experience can form an effective data collection and learning
cycle~\cite{pistar06,chi0,demavla}. Inspired by these systems, ClothDojo trains a
closed-loop data-generation policy to collect trajectories from new states.
We first train the policy on human demonstrations with supervised fine-tuning
(SFT), then perform batched rollouts. Task metrics automatically retain
successful trajectories. For failed executions, a human selects states and
provides human-in-the-loop (HIL) corrections to update the policy. Simulation also allows normalized object
coordinate space (NOCS) to be encoded as the garment texture. Through the
standard image interface of a vision-language-action
model (VLA), these coordinates provide canonical surface-position cues for
learning cloth interaction. Successful trajectories are then re-rendered with
randomized appearances to train downstream RGB policies without NOCS.
Human effort is thus concentrated on
demonstrations and necessary corrections, while the data-generation policy
scales trajectory collection through autonomous rollouts.

As shown in Fig.~\ref{fig:framework}, ClothDojo connects learned trajectory
generation, data expansion, and closed-loop evaluation through dynamic
interaction between bimanual robots and cloth. Our contributions are:

\begin{enumerate}
  \item \textbf{A simulation and evaluation platform.} A simulation platform
  for continuous bimanual garment manipulation, with shared interfaces for human
  teleoperation, policy rollouts, and closed-loop evaluation.
  \item \textbf{Learned data generation.} A data-generation pipeline that learns
  from human demonstrations and corrections, uses NOCS through a VLA's image
  interface, and autonomously collects trajectories without task-specific primitives.
  \item \textbf{Trajectory data and training value.} Nearly 8,000 successful
  trajectories across hundreds of garments. Under the same human supervision
  budget, RGB policies trained on generated data outperform those trained
  directly on human data across all evaluated garment and appearance conditions.
\end{enumerate}

\begin{figure*}[t]
  \centering
  \includegraphics[width=\textwidth]{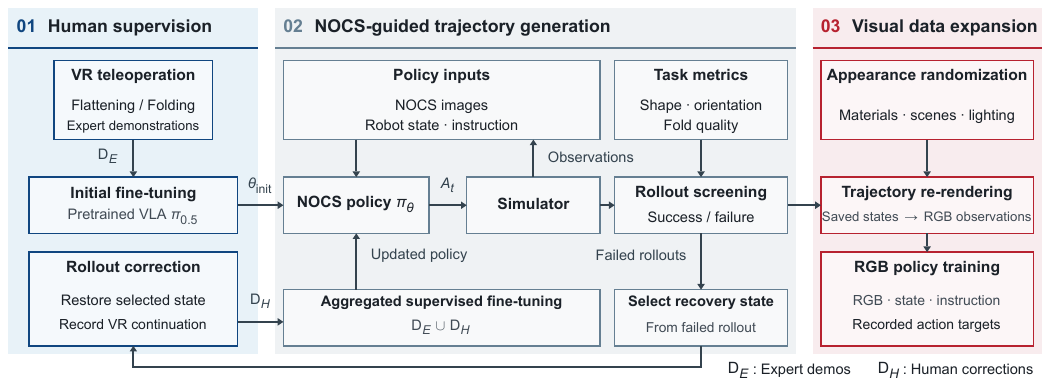}
  \caption[Overview of the NOCS-guided data-generation pipeline.]{\input{figures/clothdojo-method-framework.caption.tex}}
  \label{fig:method-framework}
\end{figure*}
\section{Related Work}
\label{sec:related-work}

\subsection{Garment manipulation policies}

Many methods define a set of task-specific primitives for garment flattening
and folding. A policy selects a primitive or predicts parameters such as grasp
locations from visual observations, while a predefined controller executes
the continuous motion~\cite{flingbot,clothmate,clothfunnels,speedfolding,unifolding}.
The policy can observe the cloth again between primitives, and some primitives
also use feedback during operations such as stretching. These action priors
reduce the learning space and have been used for flattening multiple garment
types and for structured folding.

Another approach learns continuous actions from data. Recent VLAs directly
predict action chunks and perform long sequences of picking up, flattening,
folding, and stacking garments on real robots~\cite{pistar06,chi0,demavla}.
These systems establish initial behavior with human demonstrations, expose
failures through autonomous rollouts, and use human corrections to cover
states encountered by the policy. This learning cycle also lets the policy
learn action ordering, bimanual coordination, and adjustments during execution.

\subsection{Simulation environments and data generation}

Existing garment simulation platforms provide diverse garment assets and
dynamic robot--cloth interaction for tasks including flattening and
folding~\cite{garmentlab,dexgarmentlab,lehome,simweaver}. Their publicly
described trajectories primarily come from task primitives, demonstration
transfer, or limited teleoperation.

Automatic trajectory expansion largely follows two approaches. Programmatic
methods construct trajectories using keypoints, task stages, or folding
patterns~\cite{foldnet,instantfold}. Demonstration transfer adapts segmented
behaviors to new assets and initial states, then executes them to obtain new
physical trajectories~\cite{softmimicgen,sim1}. Existing
large-scale results mainly concern folding tasks with well-defined steps.
For severely crumpled garments, however, the course of flattening changes with
the cloth state, making fixed procedures and segmented demonstrations
difficult to reuse directly.

ClothDojo brings the demonstration--rollout--correction cycle used in
real-robot garment manipulation to simulation trajectory generation. The
data-generation policy learns from human interaction and then runs autonomously
on diverse garments and complex initial states; Fig.~\ref{fig:folding-trajectories}
shows examples of the resulting behavior.

\section{Method}
\label{sec:method}

ClothDojo separates physical trajectory generation from RGB training-data
construction (Fig.~\ref{fig:method-framework}). We first train NOCS-conditioned
data-generation policies from human demonstrations and corrections, then deploy
them on a broader set of garments and initial states. Task metrics automatically
retain successful rollouts. We sample one appearance for each retained
trajectory and replay its states to render RGB observations for downstream
policy training.

\subsection{Interaction environment and tasks}

ClothDojo combines MuJoCo robot dynamics~\cite{todorov2012mujoco} with Newton's
SolverStyle3D cloth solver~\cite{newton} to support continuous bimanual
control. Virtual reality (VR) teleoperation and policy execution use the same
physics, so human
demonstrations, corrections, and autonomous rollouts share the same
robot--cloth interaction.

\begin{figure}[t]
  \centering
  \includegraphics[width=\linewidth]{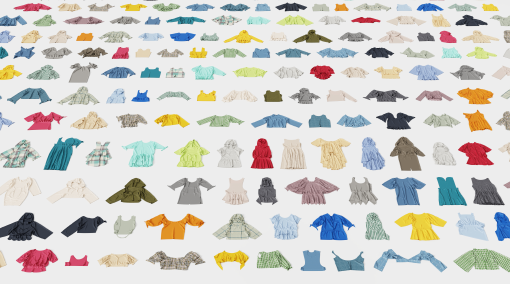}
  \caption{\textbf{The core garment pool spans diverse geometries.}
  The figure shows most of the 500 assets, covering different garment lengths,
  widths, sleeve types, necklines, and asymmetric cuts. All garments are
  naturally settled and shown at their relative physical sizes.}
  \label{fig:garment-assets}
\end{figure}

We convert more than 10,000 garments from the large-scale parametric
GarmentCodeData dataset~\cite{GarmentCodeData} into simulation assets and
select 500 structurally diverse garments as a core pool for trajectory
generation and evaluation. Human demonstrations and corrections are not
restricted to this pool. As shown in
Fig.~\ref{fig:garment-assets}, this set spans large variations in garment
length, body width, sleeve length, and sleeve width, and includes sleeveless,
hooded, low-cut, open-back, asymmetric, and unconventional designs.

\textbf{Flattening} starts from a crumpled state produced by randomly hanging,
releasing, and settling a garment. The goal is a naturally flat, face-up
configuration. All four cardinal orientations on the table are accepted, and its
position within the tabletop is unconstrained. \textbf{Folding} starts from a
preconstructed state that satisfies the flattening criterion. Depending on data
availability, these states come from human-demonstration endpoints, successful
rollouts from the data-generation policy, or standard flattened configurations. The goal
is a compact, flat rectangular bundle with little protruding material. We
prescribe no aspect ratio, crease pattern, or action order. Initial states for
both tasks are saved in advance so that policies can run from identical states;
the quantitative success criteria are given in Sec.~\ref{sec:success-criteria}.

\subsection{NOCS-conditioned data-generation policies}

We encode normalized object coordinate space
(NOCS)~\cite{wang2019nocs} as the garment texture to provide a consistent
surface-position reference as the cloth deforms. Let $v_i$ denote a vertex
of the canonical mesh, with component-wise bounds $v_{\min}$ and $v_{\max}$.
Its normalized coordinate is
\begin{equation}
  u_i=\frac{v_i-v_{\min}}{v_{\max}-v_{\min}},
  \label{eq:nocs-color}
\end{equation}
whose three components form the RGB channels. Each color is assigned in the
canonical configuration and remains attached to the corresponding vertex as
the garment deforms. Similar relative positions across garments therefore
receive similar colors. This exposes canonical surface-position cues through
the model's existing image interface. The policy still observes only visible
surfaces, and matching colors do not imply exact semantic correspondence
across garments.

We use the pretrained $\pi_{0.5}$ model~\cite{pi05} as the base and separately
fine-tune data-generation policies for flattening and folding. Each policy
receives
NOCS images through the model's original visual interface, together with the
robot state and task instruction, and predicts joint targets and gripper
commands for both arms. During execution, the policy applies only a prefix of
each predicted action sequence and then replans from the latest observation,
producing a closed-loop trajectory.

\subsection{Learning from demonstrations and corrections}
\label{sec:experience-corrections}

We collect separate flattening and folding demonstrations through VR
teleoperation. To collect a folding demonstration, we restore the final state
of its corresponding flattening demonstration and continue from that state.
The demonstrations for each task are recorded separately and used to fine-tune the respective
data-generation policies with the base model's action-prediction objective.
Sec.~\ref{sec:trajectory-agent-eval}
reports the garment coverage and number of demonstrations.

To cover states visited by the data-generation policies, we collect one round
of human corrections in a DAgger-like manner~\cite{DAgger}. For each task, we
first generate a candidate pool of 800 autonomous rollouts. A human collector
selects executions with action loops or substantial deterioration of the cloth
state, chooses an intervention point, and restores the corresponding simulation
state. The collector then uses VR to complete the task, producing a correction
trajectory $\mathcal D_H$. We combine these corrections with the original
demonstrations $\mathcal D_E$ and update the policy using the same supervised
objective. All supervised actions are provided by humans; autonomous rollouts
only supply the starting states for correction.

To add training examples with exchanged arm roles, we apply bilateral
morphological symmetry augmentation~\cite{GR-RL} to every human trajectory.
At each frame, we reflect the garment state and transform the robot states and
actions while exchanging the two arms and grippers. We then re-render NOCS
observations by replaying the transformed states. Each human trajectory yields
one mirrored sample, which is not counted as an additional human demonstration.

\subsection{RGB data from successful trajectories}

After training, we freeze the data-generation policies and generate trajectories
autonomously. For flattening, we select 400 garments and generate 20 random
initial states per garment. We execute the policy on these states, truncate a
trajectory as soon as it satisfies the success criterion in
Sec.~\ref{sec:success-criteria}, and retain at most ten successful trajectories
per garment. For folding, we sample successful terminal states from the
generated flattening trajectories and apply the four cardinal rotations to
construct a pool of 20 initial states per garment. We then use the same success
filtering and truncation procedure.

On a single NVIDIA GeForce RTX 5090 D, our implementation executes
approximately 1,152 trajectories per GPU-day, excluding RGB re-rendering.

For each successful trajectory, we sample from 34 base garment materials with
39 color variants, 17 table materials, 12 floor materials, and eight
background scenes. Light position, intensity, and color temperature are drawn
from continuous ranges. We sample one appearance per physical trajectory and
keep it fixed throughout the trajectory. We replay the saved states to render
one RGB version while preserving the robot and garment motion, robot states,
and action targets. The resulting data train
downstream policies that receive RGB observations, robot states, and task
instructions without NOCS. The appearance subsets used for training and
evaluation are specified in the experimental setup.

\section{Experiments}
\label{sec:experiments}

\subsection{Metrics and success criteria}
\label{sec:success-criteria}

\textbf{Flattening.} We align the current garment with a naturally laid-flat
reference of the same garment using translation-invariant planar Procrustes
alignment and normalize distances by the diagonal of the reference XY bounding
box. The mean three-dimensional distance between corresponding vertices after
alignment defines the deformation error $d_{\mathrm{def}}$. The mean
reference-vertex displacement induced by the difference between the fitted
rotation and the nearest valid cardinal orientation defines the orientation
error $d_{\mathrm{rot}}$; all four cardinal orientations are accepted. Using 890
expert terminal states from 182 garments, weighted equally by garment, we set
median-and-MAD upper bounds of 0.088 and 0.043. A rollout succeeds when neither
error exceeds its respective threshold for 1 s.

\textbf{Folding.} Because the task admits multiple valid outcomes, we do not
match the result to a single template. Instead, the fold-quality score $Q$
combines area compression, overall flatness, top-surface rectangularity,
three-dimensional material overlap, and proximity to the table. We fit a fixed
linear score using expert terminal states and successful HIL states as positive
examples and unfinished original rollouts as negative examples, splitting both
training and validation data by garment. The score reaches 85.4\% balanced
accuracy under garment-split cross-validation. We calibrate a threshold of
0.142 from the cross-validation predictions; a rollout succeeds when
$Q\geq0.142$ for 3 s. Score fitting and threshold calibration exclude
Unseen garments in the RGB policy evaluation.

Task-specific evaluations use 50 Hz robot control, with time limits of 60 s for
flattening and 40 s for folding. Throughout this section, arrows indicate the
preferred direction of a metric; for comparisons among like policies, bold
marks the best result for each task and evaluation condition.

\subsection{Can the data-generation policy scale trajectory collection?}
\label{sec:trajectory-agent-eval}

We test whether NOCS, mirror augmentation, and HIL progressively improve the
policy's ability to generate successful trajectories. Each policy first learns
from expert demonstrations; HIL corrections then cover states encountered
during autonomous execution. Table~\ref{tab:human-training-data} summarizes the
two sources of human data and their garment coverage. Folding demonstrations
start from the terminal states of their corresponding flattening
demonstrations. Because these states are relatively controlled, we collect
fewer demonstrations per garment.
All trajectories are cleaned before training; the table also reports their
sizes after the mirror augmentation in Sec.~\ref{sec:experience-corrections}.

\begin{table}[!ht]
  \centering
  \caption{Human training data for the data-generation policies.}
  \label{tab:human-training-data}
  \footnotesize
  \setlength{\tabcolsep}{4.0pt}
  \begin{tabular}{@{}lcc@{}}
    \toprule
    \smash{\raisebox{-0.9ex}{Task}} & Expert & HIL corrections \\
    & raw $\rightarrow$ aug. / garments & raw $\rightarrow$ aug. / garments \\
    \midrule
    Flatten & 890 $\rightarrow$ 1,780 / 182 & 201 $\rightarrow$ 402 / 88 \\
    Fold & 537 $\rightarrow$ 1,074 / 156 & 213 $\rightarrow$ 426 / 125 \\
    \bottomrule
  \end{tabular}
\end{table}

Starting from White, which uses expert demonstrations with a uniform white
garment appearance, we successively add NOCS, mirror augmentation, and HIL
corrections to form four cumulative configurations. Except for these cumulative
changes, all configurations use the same training hyperparameters, train for
three epochs on their respective datasets, and evaluate the final checkpoint.
All data-generation policies are initialized from the pretrained $\pi_{0.5}$
model and trained separately for flattening and folding. For each task, we
evaluate 200 garments from four saved initial states,
for 800 closed-loop rollouts in total. All configurations share the same initial
states and execution settings. Seen garments occur in $\mathcal D_E$
(or $\mathcal D_H$ for folding), whereas Unseen garments occur in neither
$\mathcal D_E$ nor $\mathcal D_H$. The flattening evaluation has 100 Seen and
100 Unseen garments; folding has 130 and 70, respectively. We
report overall and split success rates, together with the successful-trajectory
quality metrics defined in Sec.~\ref{sec:success-criteria}. The Expert rows use terminal states that
pass the success criteria as quality references and are excluded from the
success-rate comparison. In Table~\ref{tab:flatten-performance}, Def. and Rot.
are garment-equal means at first confirmed success.

\begin{table}[!ht]
  \centering
  \caption{Data-generation-policy performance on flattening.}
  \label{tab:flatten-performance}
  \footnotesize
  \setlength{\tabcolsep}{3.1pt}
  \begin{tabular}{@{}lccccc@{}}
    \toprule
    Method & SR (\%) $\uparrow$ & Seen & Unseen &
      Def. $\downarrow$ & Rot. $\downarrow$ \\
    \midrule
    White & 24.8 & 26.3 & 23.3 & 0.081 & 0.024 \\
    + NOCS & 62.9 & 66.3 & 59.5 & 0.080 & 0.024 \\
    + Mirror & 69.6 & 76.0 & 63.3 & 0.080 & 0.025 \\
    + HIL & \textbf{76.5} & \textbf{82.3} & \textbf{70.8} &
      0.080 & 0.024 \\
    \midrule
    Expert & -- & -- & -- & 0.068 & 0.019 \\
    \bottomrule
  \end{tabular}
\end{table}

\begin{table}[!ht]
  \centering
  \caption{Data-generation-policy performance on folding.}
  \label{tab:fold-performance}
  \footnotesize
  \setlength{\tabcolsep}{5.0pt}
  \begin{tabular}{@{}lcccc@{}}
    \toprule
    Method & SR (\%) $\uparrow$ & Seen & Unseen & $Q$ $\uparrow$ \\
    \midrule
    White & 55.1 & 61.0 & 44.3 & 0.809 \\
    + NOCS & 57.6 & 61.2 & 51.1 & 0.779 \\
    + Mirror & 60.5 & 64.2 & 53.6 & 0.846 \\
    + HIL & \textbf{70.0} & \textbf{73.7} & \textbf{63.2} & 0.825 \\
    \midrule
    Expert & -- & -- & -- & 1.037 \\
    \bottomrule
  \end{tabular}
\end{table}

NOCS provides the largest gain for flattening, increasing overall success by
38.1 percentage points. Its overall effect on folding is smaller: performance
is nearly unchanged on Seen garments but improves by 6.8 points on Unseen
garments. This contrast suggests that canonical position cues matter most when
severe deformation makes garment parts difficult to identify, while also
supporting transfer of learned manipulation across garment structures.

Mirror augmentation further improves both tasks without adding human
trajectories, by 6.7 points for flattening and 2.9 points for folding. The larger
flattening gain is consistent with its arbitrary initial orientations and
flexible division of labor between the arms, indicating that known bilateral
symmetry can expand the execution patterns covered by limited human data.

HIL further improves both tasks and provides the largest single-step gain for
folding. Unlike NOCS and mirror augmentation, HIL adds corrections from failure
states actually visited by the policy. With HIL, success on Unseen garments
reaches 70.8\% for flattening and 63.2\% for folding, enabling the policy to
expand successful trajectories beyond the human data.

\begin{figure}[!ht]
  \centering
  \includegraphics[width=\columnwidth]{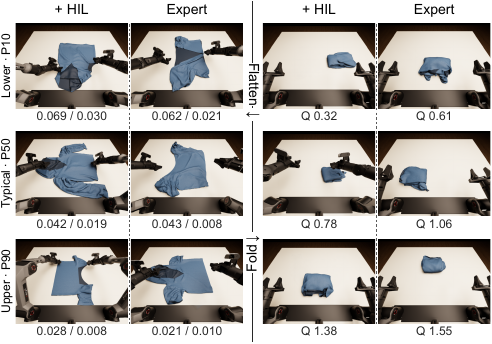}
  \caption{\textbf{The data-generation policy produces successful terminal
  states with visual quality close to human demonstrations.} Rows show selected
  released states at P10, P50, and P90 of each source's eligible subset, ranked jointly
  by deformation and orientation errors for flattening, and by $Q$ for folding. The two
  values below each flattening image are $d_{\mathrm{def}}$ and
  $d_{\mathrm{rot}}$, respectively (lower is better); folding images report $Q$
  (higher is better). Flattening images use later released states, whereas
  Table~\ref{tab:flatten-performance} measures first confirmed success.}
  \label{fig:quality-boundaries}
\end{figure}

Higher success rates do not coincide with uniformly better success-state
metrics. Def. and Rot. remain stable across flattening configurations, whereas
$Q$ fluctuates because each value is measured on a different successful subset.
Human demonstrations have finer terminal states on average.
Fig.~\ref{fig:quality-boundaries} compares their visual quality with successful
policy rollouts at selected quality percentiles.

\subsection{What is the value of automatically generated data?}
\label{sec:generated-data-value}

We test whether autonomously generated trajectories improve RGB policy learning
without additional human supervision. Starting from the same pretrained
$\pi_{0.5}$ weights, we train separate RGB policies for flattening and folding
on three data sources. \textbf{Human} contains the original demonstrations and
corrections, whereas \textbf{Generated} contains autonomous trajectories
retained by the success criteria. Their garment coverage partially overlaps,
and \textbf{Combined} pools both datasets. Human and Generated use the same RGB
re-rendering procedure. All three conditions draw on the same original human
data, and autonomous generation adds no human action labels. Each policy uses
the same training hyperparameters, trains for three epochs on its respective
dataset, and is evaluated at the final checkpoint.
Table~\ref{tab:rgb-policy-training-data} summarizes the number of trajectories
and garments in each training set.

\begin{table}[!ht]
  \centering
  \caption{RGB-policy training data (trajectories / garments).}
  \label{tab:rgb-policy-training-data}
  \footnotesize
  \setlength{\tabcolsep}{4.0pt}
  \begin{tabular}{@{}lccc@{}}
    \toprule
    Task & Human & Generated & Combined \\
    \midrule
    Flatten & 1,091 / 217 & 3,992 / 400 & 5,083 / 470 \\
    Fold & 750 / 242 & 3,856 / 399 & 4,606 / 458 \\
    \bottomrule
  \end{tabular}
\end{table}

We evaluate garment and appearance generalization on a shared test set. The
Seen split contains 100 garments represented in all three training sets, while
the Unseen split contains another 100 garments absent from all three. Of these
Unseen garments, 73 come from the core 500-garment pool and 27 from the larger
converted asset pool. For each garment, we evaluate four fixed held-out initial
states under both Seen and Unseen appearances. Seen appearances are sampled from the training
randomization distributions; Unseen appearances use garment textures, support
surfaces, backgrounds, and lighting conditions withheld from training. All
policies share the garments, initial states, and appearance settings. Each of
the four garment--appearance combinations therefore contains 400 closed-loop
rollouts; Avg. is their unweighted mean.

\begin{table}[!t]
  \centering
  \caption{RGB-policy success rates (\%) by training set.}
  \label{tab:generated-data-value}
  \footnotesize
  \setlength{\tabcolsep}{2.65pt}
  \renewcommand{\arraystretch}{1.10}
  \begin{tabular}{@{}Tlccccc@{}}
    \toprule
    \multirow[c]{2}{=}{Task} & \multirow[c]{2}{*}{Training set}
      & \multicolumn{2}{c}{Seen garments}
      & \multicolumn{2}{c}{Unseen garments}
      & \multirow[c]{2}{*}{Avg.} \\
    \cmidrule(lr){3-4}\cmidrule(lr){5-6}
      & & \multicolumn{1}{c}{S app.} & \multicolumn{1}{c}{U app.}
      & \multicolumn{1}{c}{S app.} & \multicolumn{1}{c}{U app.} & \\
    \midrule
    \multirow[c]{3}{=}{Flatten} & Human
      & 33.3 & 30.5 & 22.3 & 19.3 & 26.3 \\
    & Generated
      & \textbf{41.3} & \textbf{33.3} & \textbf{23.5} & \textbf{23.8} & \textbf{30.4} \\
    & Combined
      & 32.8 & 28.0 & 20.3 & 19.5 & 25.1 \\
    \midrule
    \multirow[c]{3}{=}{Fold} & Human
      & 68.0 & 61.8 & 40.0 & 37.8 & 51.9 \\
    & Generated
      & \textbf{72.3} & 69.0 & \textbf{47.8} & \textbf{47.0} & \textbf{59.0} \\
    & Combined
      & 71.0 & \textbf{69.8} & 43.0 & 46.0 & 57.4 \\
    \bottomrule
    \multicolumn{7}{@{}l@{}}{\scriptsize S app. and U app. denote Seen and Unseen appearances, respectively.} \\
  \end{tabular}
\end{table}

Although Generated trajectories end in less refined states on average than
Human trajectories, the RGB policies trained on Generated achieve higher
success rates in every evaluation condition in
Table~\ref{tab:generated-data-value}. Autonomous generation therefore broadens
the training experience available to downstream policies.

Combined contains the most trajectories and garments, yet provides no further
improvement. The benefit of Generated therefore cannot be explained by data scale
and coverage alone: the distribution induced by autonomous rollouts and
success filtering may also matter, and simply pooling human and generated
trajectories does not automatically exploit their complementarity. How to
combine the two sources effectively remains an open question.

Across the two generalization axes, Unseen garments cause a larger performance
drop than appearance shifts, consistently across all three training sets and
both tasks. Generalizing to new garment structures and deformation patterns
therefore remains the primary bottleneck for current RGB policies.

\subsection{Comparison of policy models}
\label{sec:policy-model-comparison}

We train four policy models on the Generated data from
Sec.~\ref{sec:generated-data-value} and compare them under the same closed-loop
protocol across tasks, garments, and appearances. Specifically, we fine-tune
$\pi_{0.5}$~\cite{pi05}, GR00T N1.7~\cite{gr00tn17}, and an internal policy
implemented with the public StarVLA framework~\cite{starvla}, and train
Diffusion Policy~\cite{diffusionpolicy} from scratch.

\begin{table}[!t]
  \centering
  \caption{RGB-policy success rates (\%) by method.}
  \label{tab:policy-model-comparison}
  \footnotesize
  \setlength{\tabcolsep}{2.65pt}
  \renewcommand{\arraystretch}{1.10}
  \begin{tabular}{@{}Tlccccc@{}}
    \toprule
    \multirow[c]{2}{=}{Task} & \multirow[c]{2}{*}{Method}
      & \multicolumn{2}{c}{Seen garments}
      & \multicolumn{2}{c}{Unseen garments}
      & \multirow[c]{2}{*}{Avg.} \\
    \cmidrule(lr){3-4}\cmidrule(lr){5-6}
      & & \multicolumn{1}{c}{S app.} & \multicolumn{1}{c}{U app.}
      & \multicolumn{1}{c}{S app.} & \multicolumn{1}{c}{U app.} & \\
    \midrule
    \multirow[c]{4}{=}{Flatten} & $\pi_{0.5}$
      & \textbf{41.3} & \textbf{33.3} & \textbf{23.5} & \textbf{23.8} & \textbf{30.4} \\
    & StarVLA
      & 25.0 & 14.3 & 12.8 & 8.3 & 15.1 \\
    & GR00T
      & 16.0 & 17.3 & 10.8 & 12.3 & 14.1 \\
    & Diffusion
      & 3.5 & 4.5 & 1.3 & 3.0 & 3.1 \\
    \midrule
    \multirow[c]{4}{=}{Fold} & $\pi_{0.5}$
      & \textbf{72.3} & \textbf{69.0} & \textbf{47.8} & \textbf{47.0} & \textbf{59.0} \\
    & StarVLA
      & 67.0 & 67.5 & 38.3 & 40.3 & 53.3 \\
    & GR00T
      & 47.0 & 47.0 & 26.5 & 27.0 & 36.9 \\
    & Diffusion
      & 16.8 & 5.0 & 10.8 & 2.0 & 8.6 \\
    \bottomrule
    \multicolumn{7}{@{}l@{}}{\scriptsize S app. and U app. denote Seen and Unseen appearances, respectively.} \\
  \end{tabular}
\end{table}

Table~\ref{tab:policy-model-comparison} shows that $\pi_{0.5}$ achieves the
highest success rate across both tasks and all evaluation conditions, whereas
Diffusion Policy achieves the lowest; StarVLA and GR00T lie between them. These
consistent gaps under a common protocol show that ClothDojo can differentiate
policy models by their closed-loop performance.

Consistent with Sec.~\ref{sec:generated-data-value}, all four models degrade on
Unseen garments. Appearance shifts, however, have model- and task-dependent
effects: they affect StarVLA mainly in flattening and Diffusion Policy mainly in
folding, while $\pi_{0.5}$ and GR00T remain comparatively stable. Garment
generalization and appearance robustness therefore capture two complementary
dimensions of policy performance.

All four policies achieve lower success rates on flattening than on folding in
our evaluation. The two tasks differ in both initial-state complexity and
behavioral structure. Flattening starts from randomly crumpled states, requiring
the policy to recognize garment structure under self-occlusion and adapt its
regrasping, unfolding, and local adjustments as the garment evolves. Folding
starts from a valid flattened state, where garment structure is clearer and the
manipulation sequence is more regular. Flattening therefore requires more
state-dependent long-horizon decisions.

\subsection{Task composition and cross-embodiment reuse}
\label{sec:additional-analyses}

As a supplementary experiment, we use the full configuration in
Sec.~\ref{sec:trajectory-agent-eval}, including NOCS, mirror augmentation, and
HIL, to study task composition and cross-embodiment reuse. For task composition,
we sample flattening and folding data equally and jointly train the policy to
predict the semantic subtask and its corresponding action under the overall
instruction ``flatten and fold the garment.'' At inference time, the policy
selects the current subtask from its observation. For cross-embodiment reuse,
we retarget the same training trajectories to UR5e using inverse kinematics and
dynamics replay, without collecting additional human trajectories.

Flatten $\rightarrow$ Fold starts from a crumpled state and runs continuously
for 120 s; Fold starts from a fixed state that satisfies the flattening criterion
and runs for 40 s. Both evaluations use the task metrics and initial-state setup
from Sec.~\ref{sec:trajectory-agent-eval}. For Flatten $\rightarrow$ Fold, the
two SRs use the same set of trials and separately measure the fraction that
reaches each criterion. Table~\ref{tab:joint-policy} reports the results.

The joint policy selects semantic subtasks from its observations and reaches
both success criteria in some executions, but it does not yet complete the full
process reliably; its performance in the standalone Fold evaluation also trails
that of the task-specific policy. Balanced joint training and semantic
supervision thus provide an initial mechanism for task composition, while task
interference and the lack of transition experience may still limit continuous
execution.

\begin{table}[!t]
  \centering
  \caption{Task composition and cross-embodiment reuse.}
  \label{tab:joint-policy}
  \footnotesize
  \setlength{\tabcolsep}{4.5pt}
  \begin{tabular}{@{}lccc@{}}
    \toprule
    \raisebox{-0.9ex}[0pt][0pt]{Robot} & \multicolumn{2}{c}{Flatten $\rightarrow$ Fold} & Fold \\
    \cmidrule(lr){2-3}\cmidrule(l){4-4}
    & Flatten SR & Fold SR & Fold SR \\
    \midrule
    Piper & 49.4 & 20.0 & 17.8 \\
    UR5e & 68.9 & 7.1 & 27.8 \\
    \bottomrule
  \end{tabular}
\end{table}

The UR5e policy, trained only on retargeted trajectories, can likewise perform
both flattening and folding. ClothDojo's data-generation process is therefore
not tied to Piper, although reuse requires the target robot's workspace and
kinematics to be compatible with the original trajectories.

\begin{figure}[!ht]
  \centering
  \includegraphics[width=\columnwidth]{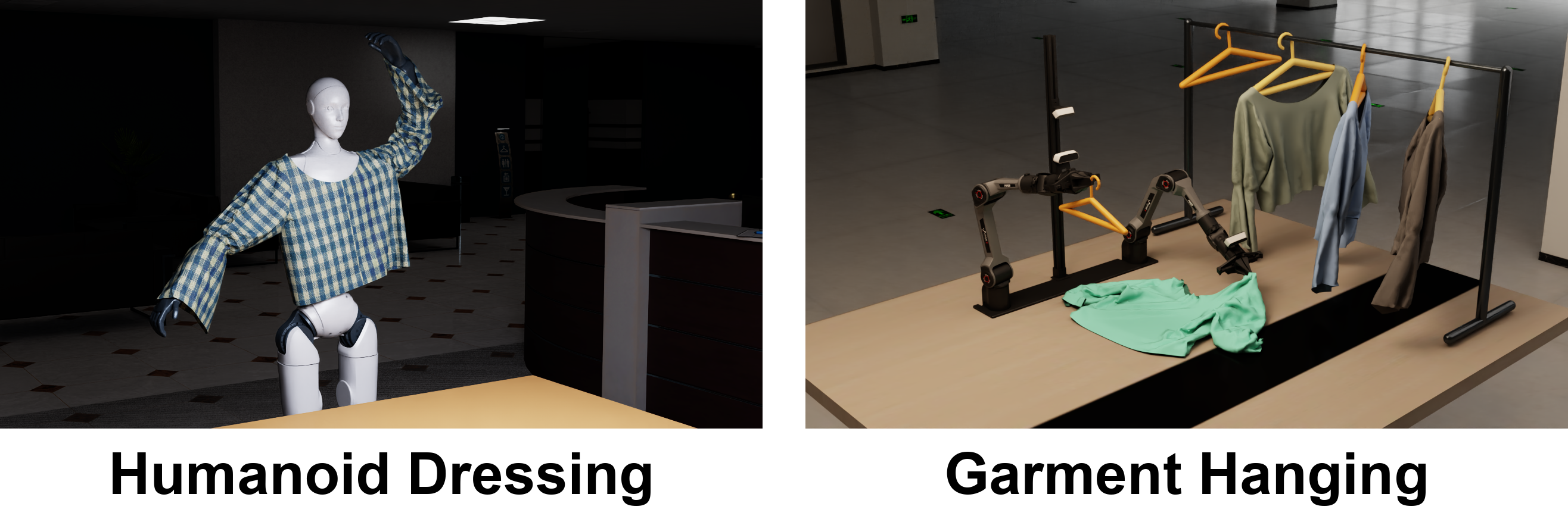}
  \caption{\textbf{Potential task extensions of ClothDojo.} Left:
  humanoid dressing. Right: bimanual garment hanging. Both settings introduce
  more complex cloth--rigid-body contacts than tabletop flattening and folding.}
  \label{fig:future-task-extensions}
\end{figure}

\section{Conclusion}
\label{sec:conclusion}

\textbf{Contributions.} ClothDojo unifies learned trajectory generation and
closed-loop evaluation for continuous bimanual garment manipulation. Its
infrastructure supports VR demonstrations, HIL corrections, batched policy
execution, success filtering, and state-based RGB re-rendering for both
flattening and folding. Using NOCS-conditioned policies, we generate nearly
8,000 successful trajectories across hundreds of garments; under our evaluation
protocol, RGB policies trained on these trajectories outperform those trained
directly on the original human trajectories. The trained checkpoints and
pipeline can continue generating trajectories for new garments and initial
states, making ClothDojo an extensible source of adaptive garment-manipulation
data rather than a fixed dataset.

\textbf{Limitations.} We validate the complete ClothDojo pipeline only in
simulation and do not evaluate the learned policies on real robots. Evaluation
at a comparable scale would require repeated physical setup and execution;
reliable transfer also requires real-to-sim calibration of cloth material and
contact parameters, robot dynamics, and low-level control---a
substantial research problem in its own right. Finally, the training
trajectories do not cover the transition from flattening to folding, and the
joint policy does not yet execute the two stages reliably in sequence.

\textbf{Future work.} The current tasks primarily involve cloth self-contact
and relatively simple cloth--rigid-body contact. We will leverage the solver's
bidirectional force coupling and watertight contact handling to extend
ClothDojo to more contact-rich tasks, including humanoid dressing and garment
hanging (Fig.~\ref{fig:future-task-extensions}).

\bibliographystyle{IEEEtran}
\IEEEtriggeratref{11}
\IEEEtriggercmd{\newpage\vspace*{-12.56pt}}
\bibliography{references}

\end{document}